\documentclass[journal]{IEEEtran}
\usepackage{epsfig}
\usepackage{graphicx}
\usepackage{amsmath}
\usepackage{amssymb}
\usepackage[ruled]{algorithm2e}
\usepackage{bbm}
\usepackage{multirow}
\usepackage{bm}
\usepackage{url}
\usepackage{xspace}
\usepackage{amsmath,amsfonts}
\usepackage{amsthm}
\usepackage{amssymb,amsopn}
\usepackage{bm} 

\newcommand{\vct}[1]{\boldsymbol{#1}} 
\newcommand{\mat}[1]{\boldsymbol{#1}} 

\newcommand{\T}{^{\textrm T}} 

\DeclareMathOperator{\argmax}{arg\,max}

\newcommand{\eat}[1]{}

\usepackage{hyperref}
\usepackage{xcolor}
\usepackage{booktabs}
\hypersetup{
    colorlinks=true,
    linkcolor=blue,
    filecolor=magenta,      
    urlcolor=blue,
}

\newcommand{\etal}{\emph{et al.}\xspace}
\newcommand{\ie}{\emph{i.e.}\xspace}
\newcommand{\eg}{\emph{e.g.}\xspace}
\newcommand{\tabincell}[2]{\begin{tabular}{@{}#1@{}}#2\end{tabular}}

\ifCLASSINFOpdf
\else
\fi
\begin{document}
%
\title{SPICE: Semantic Pseudo-Labeling for \\ Image Clustering}
%
%
%

\author{Chuang Niu,~\IEEEmembership{Member,~IEEE,}
        Hongming Shan,~\IEEEmembership{Member,~IEEE,}
        and~Ge~Wang,~\IEEEmembership{Fellow,~IEEE}

\thanks{ C. Niu and G. wang are with Department of Biomedical Engineering, Center for Biotechnology and Interdisciplinary Studies, Rensselaer Polytechnic Institute, Troy, NY USA, 12180. E-mail: niuc@rpi.edu; wangg6@rpi.edu.}
\thanks{H. Shan is with the Institute of Science and Technology for Brain-inspired Intelligence and MOE Frontiers Center
for Brain Science, Fudan University, Shanghai 200433, China, and also with the Shanghai Center for Brain Science and Brain-Inspired
Technology, Shanghai 201210, China. Email: hmshan@fudan.edu.cn.}
\thanks{Manuscript received xx xx, 2022; revised xx xx, 2022.}}

%
%

\markboth{A Manuscript}%
{Shell \MakeLowercase{\textit{et al.}}: Bare Demo of IEEEtran.cls for IEEE Journals}
%



\maketitle

\begin{abstract}
    The similarity among samples and the discrepancy among clusters are two crucial aspects of image clustering. However, current deep clustering methods suffer from inaccurate estimation of either feature similarity or semantic discrepancy.
    In this paper, we present a Semantic Pseudo-labeling-based Image ClustEring (SPICE) framework, which divides the clustering network into a feature model for measuring the instance-level similarity and a clustering head for identifying the cluster-level discrepancy.
    We design two semantics-aware pseudo-labeling algorithms, prototype pseudo-labeling and reliable pseudo-labeling, which enable accurate and reliable self-supervision over clustering.
    Without using any ground-truth label, we optimize the clustering network in three stages: 1) train the feature model through contrastive learning to measure the instance similarity;
    2) train the clustering head with the prototype pseudo-labeling algorithm to identify cluster semantics;
    and 3) jointly train the feature model and clustering head with the reliable pseudo-labeling algorithm to improve the clustering performance.
    Extensive experimental results demonstrate that SPICE achieves significant improvements ($\sim$10\%) over existing methods and establishes the new state-of-the-art clustering results on six image benchmark datasets in terms of three popular metrics. 
    Importantly, SPICE significantly reduces the gap between unsupervised and fully-supervised classification; \eg there is only 2\% (91.8\% vs 93.8\%) accuracy difference on CIFAR-10. 
    Our code is made publically available at~\url{https://github.com/niuchuangnn/SPICE}.
\end{abstract}


%
\IEEEpeerreviewmaketitle

\section{Introduction}
%
%
%
%
\IEEEPARstart{I}{mage} clustering aims to group images into different meaningful clusters without human annotations, and is an essential task in unsupervised learning with the applications in many areas. At the core of image clustering are the measurements of the similarity among samples (images) and the discrepancy among semantic clusters. 
Recently, deep learning based clustering methods  have achieved great progress thanks to the strong representation capability of deep neural networks. 

\begin{figure}[bt!]
    \centering
    \includegraphics[width=1\linewidth]{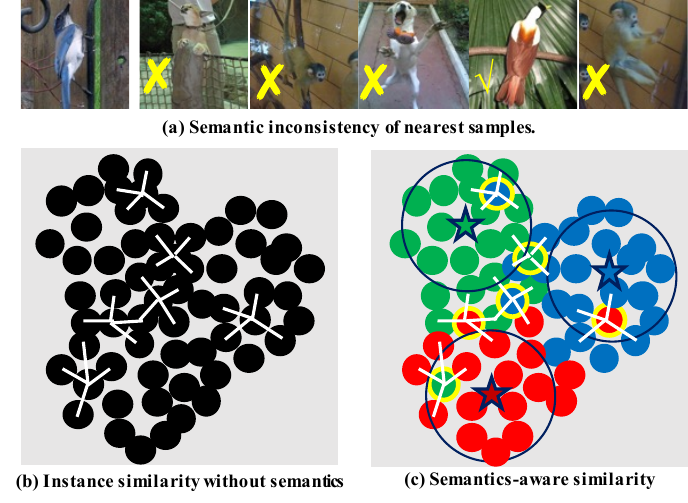}
    \caption{Semantic relevance in the feature space. (a) Neighboring samples of different semantics, where the first image is the query image and the other images are nearest images with closest features provided by SCAN~\cite{scan}. (b) Instance similarity without semantics, where each point denotes a sample in the feature space, and white lines indicate similar samples. (c) Semantics-aware similarity, where different colors denote different semantic clusters, stars denote cluster centers, the points within large circles are similar to the cluster centers, and the points with yellow circles are semantically inconsistent to neighbor samples.  }
    \label{fig_general}
\end{figure}

Initially, by combining autoencoders with clustering algorithms, some deep clustering methods were proposed to learn representation features and perform clustering simultaneously and alternatively~\cite{Xie2016, LI2018161, DCN2016, DeepCluster2017, Zhang_2019_CVPR, DEPICT2017, VaDE2017, GMVAE, DASC2018}, achieving better results than the traditional methods. Due to the overestimation of low-level features, these autoencoder-based methods hardly capture discriminitative features of complex images. Thus, a number of methods were proposed to learn discriminitative label features under various constraints~\cite{DAIC2017,IIC2019,Wu_2019_ICCV,gatcluster,gatcluster, cc, idfd}. However, these methods have limited performance when directly using the label features to measure the similarity among samples. This is because the category-level features lose too much instance-level information to accurately measure the instance similarity.
Very recently, Van Gansbeke~\etal~\cite{scan} proposed to leverage the embedding features of unsupervised representation learning model to search for similar samples across the whole dataset, and then encourage a clustering model to output the same labels for similar instances, which further improved the clustering performance.
Considering the imperfect embedding features, the local nearest samples in the embedding space do not always have the same semantics especially when the samples lie around the borderlines between different clusters as shown in Fig.~\ref{fig_general}(a), which may compromise the performance.
Essentially, SCAN only utilizes the instance similarity for training the clustering model without explicitly exploring the semantic discrepancy between clusters, as shown in Fig.~\ref{fig_general}(b), so that it cannot identify the semantically inconsistent samples.

\begin{figure*}[bt!]
    \centering
    \includegraphics[width=1\textwidth]{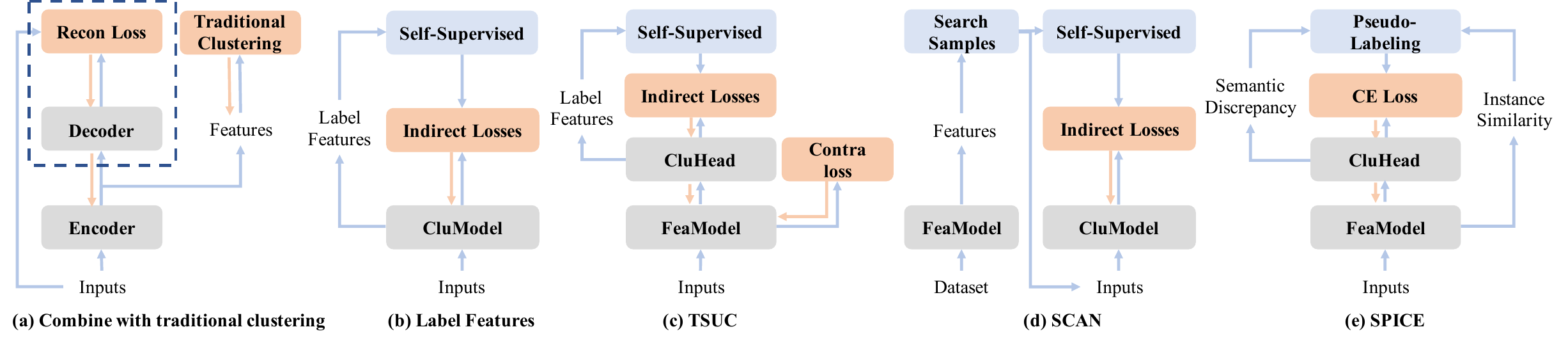}
    \caption{Training framework of different  deep clustering methods. (a) The initial deep clustering methods that combine traditional clustering algorithms with the deep neural networks, most of them combine with the autoencoders and some combine with an encoder only. (b) The label feature based methods that directly map images to cluster labels, where self-supervision is calculated based on the label features only. (c) A two-stage unsupervised classification method that first trains the feature model and then trains the whole classification work with a label feature based self-supervision and an embedding feature based contrastive loss simultaneously and separately. (d) The SCAN learning framework that constrains the similar samples in the embedding space having the same cluster label. (e) The proposed SPICE framework that synergizes both the similarity among samples and the discrepancy between clusters for training a clustering network through semantics-aware pseudo-labeling.}
    \label{fig_arcs}
\end{figure*}

To this end,  we propose a Semantic Pseudo-labeling-based Image ClustEring (SPICE) framework that synergizes the similarity among instances and semantic discrepancy between clusters to generate accurate and reliable self-supervision over clustering.
In SPICE, the clustering network is divided into two parts: a feature model and a subsequent clustering head, which is exactly a traditional classification network.
To effectively measure the instance similarity and the cluster discrepancy, we split the training process into three stages: 1) training the feature model; 2) training the clustering head; and 3) jointly training the feature model and clustering head. We highlight that there is no any annotations throughout the training process.
More specifically, in the first stage we adopt the self-supervised contrastive learning paradigm to train the feature model, which can accurately measure the similarity among instance samples. In the second stage, we  propose a prototype pseudo-labeling algorithm to train the clustering head in the expectation-maximization (EM) framework with the feature model fixed, which can take into account both the instance similarity and the semantic discrepancy for clustering. In the final stage, we propose a reliable pseudo-labeling algorithm to jointly optimize the feature model and the clustering head, which can boost both the clustering performance and the representation ability.

Compared with the instance similarity based method~\cite{scan}, the proposed prototype pseudo-labeling algorithm can leverage the predicted cluster labels, obtained from the clustering head, to identify cluster prototypes in the feature space, and then assign each prototype label to its neighbor samples for training the clustering head alternatively. Thus, the inconsistency among borderline samples can be avoided when the prototypes are well captured, as shown in Fig.~\ref{fig_general}(b).
On the other hand, given the predicted labels, the reliable pseudo-labeling algorithm can identify unreliable samples in the embedding space, as the yellow circles in Fig.~\ref{fig_general}(c), which will be filtered out during joint training. Therefore, the proposed SPICE can generate more accurate and reliable supervision by synergizing the similarity and discrepancy.

Our main contributions are summarized as follows.
\begin{enumerate}
\item  We propose a novel SPICE framework for image clustering, which can generate accurate and reliable self-supervision over clustering by synergizing the similarity among samples and the discrepancy between clusters.
\item We use the divide-and-conquer strategy to divide the clustering network into a feature model and a clustering head, and gradually train the feature model, clustering head, and both of them jointly, without using any annotations.
\item  We design a prototype pseudo-labeling algorithm to identity prototypes for training the clustering head in an EM framework, which can reduce the semantic inconsistency of the samples around borderlines.
\item  We design a reliable pseudo-labeling algorithm to select reliable samples for jointly training the feature model and clustering head, which effectively improves the clustering performance.
\item  Extensive experimental results demonstrate that SPICE outperforms the state-of-the-art clustering methods on common image clustering benchmarks by a large margin ($\sim$10\%), closing the gap between unsupervised and supervised classification (down to $\sim$2\% on CIFAR10).
\end{enumerate}

\section{Related work}

In this section, we first analyze the deep image clustering methods systematically, and then briefly review the related unsupervised representation learning and semi-supervised classification methods.

\subsection{Deep Clustering}
\label{sec_udc}
Deep clustering methods have shown significant superiority over traditional clustering algorithms, especially in computer vision.
In a data-driven fashion, deep clustering can effectively utilize the representation ability of deep neural networks.
Initially, some methods were proposed to combine deep neural networks with traditional clustering algorithms, as shown in Fig.~\ref{fig_arcs}(a).
Most of these clustering methods combine the stacked auto-encoders (SAE)~\cite{SDAE2010} with the traditional clustering algorithms, such as k-means~\cite{DEN2014, Xie2016, DCN2016, DMC2017, LI2018161}, Gaussian mixture model~\cite{DeepCluster2017, VaDE2017, GMVAE}, spectral clustering~\cite{DSCN2017}, subspace clustering~\cite{DASC2018, Zhang_2019_CVPR}, and relative entropy constraint~\cite{DEPICT2017}. However, since the pixel-wise reconstruction loss of SAE tends to over-emphasize low-level features, these methods have inferior performance in clustering images of complex contents due to the lack of object-level semantics.
Instead of using SAE, Yang~\etal~\cite{Yang2016Joint} alternately perform the agglomerative clustering~\cite{Agglomerative} and train the representation model by enforcing the samples within a selected cluster and its nearest cluster having similar features while pushing away the selected cluster from its other neighbor clusters. However, the performance of JULE can be compromised by the errors accumulated during the alternation, and their successes in online scenarios are limited as they need to perform clustering on the entire dataset.

Recently, novel methods emerged that directly learn to map images into label features, which are used as the representation features during training and as the one-hot encoded cluster indices during testing~\cite{pami-c2, DAIC2017, dsec, Wu_2019_ICCV, IIC2019, Huang_2020_CVPR, gatcluster, scan, cc, pami-c1, idfd}, as shown in Fig.~\ref{fig_arcs}(b). Actually, these methods aim to train the classification model in the unsupervised setting while using multiple indirect loss functions, such as sample relations~\cite{DAIC2017}, invariant information~\cite{IIC2019, cc}, mutual information~\cite{Wu_2019_ICCV}, partition confidence maximisation~\cite{Huang_2020_CVPR}, attention~\cite{gatcluster}, and entropy~\cite{gatcluster, Huang_2020_CVPR, scan, cc}. 
Gupta~\etal~\cite{Gupta2020Unsupervised} proposed to train an ensemble of deep networks and select the predicted labels that a large number of models agree on as the high-quality labels, which are then used to train a ladder network~\cite{ladder} in a semi-supervised learning mode.
However, the performance of these methods may be sub-optimal when using such label features to compute the similarity and discrepancy between samples, as the category-level label features can hardly reflect the relations of instance-level samples accurately.

To improve the representation learning ability, a two-stage unsupervised classification method (TSUC)~\cite{tsuc} was proposed. In the first stage, the unsupervised representation learning model was trained for initialization. In the second stage, a mutual information loss~\cite{Wu_2019_ICCV} based on the label features and a contrastive loss based on the embedding features were simultaneously optimized for training the whole model, as shown in Fig.~\ref{fig_arcs}(c).
Although the better initialization and contrastive loss can help learn representation features, the supervision based on the similarity and the discrepancy are computed independently, and the end-to-end training without accurate discriminative supervision will even harm the representation features as analyzed in Section~\ref{sec_abl}.
In contrast, Van Gansbeke~\etal~\cite{scan} proposed a method called SCAN to use embedding features of the representation learning model for computing the instance similarity, based on which the label features are learned by encouraging similar samples to have the same label, as shown in Fig.~\ref{fig_arcs}(d).
Sharing the same idea with SCAN, NNM~\cite{Dang_2021_CVPR} was proposed to enhance SCAN by searching similar samples on both the entire dataset and the sub-dataset.
However, since the embedding features are not perfect, similar instances do not always have the same semantics especially when the samples lie near the borderlines of different clusters. Therefore, only using the instance similarity and ignoring the semantic discrepancy between clusters to guide model training may limit the clustering performance.
Recently, Park~\etal~\cite{Park_2021_CVPR} proposed an add-on module for improving the off-the-shelf unsupervised clustering method based on the semi-supervised learning~\cite{mixmatch} and label smoothing techniques~\cite{Li2020DivideMix, pmlr-v119-lukasik20a}.

Based on the comprehensive analysis of existing deep cluttering methods, we present a new framework for image clustering, as shown in Fig.~\ref{fig_arcs}(e), which can accurately measure the similarity among samples and the discrepancy between clusters for the model training, and effectively reduce the semantic inconsistency for similar samples.
In addition to the image clustering methods mentioned above, there are also  promising deep learning-based clustering methods focusing on special~\cite{TIP4, TIP5} or multi-view clustering datasets~\cite{TIP2, TIP3}.

\begin{figure*}[bt!]
    \centering
    \includegraphics[width=0.85\textwidth]{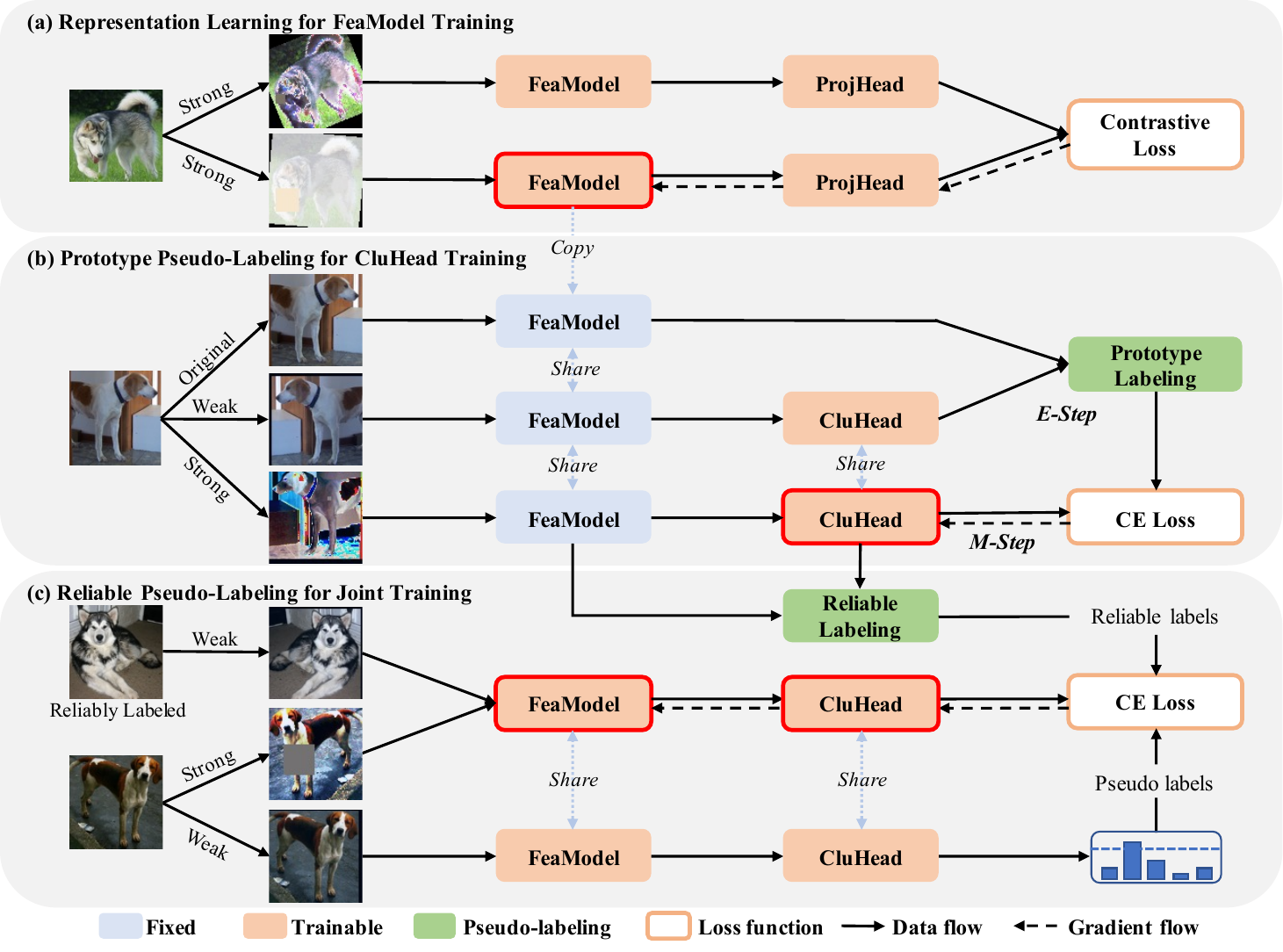}
    \caption{Illustration of the SPICE framework. (a) Train the feature model with the contrastive learning based unsupervised representation learning. (b) Train the clustering head via the prototype pseudo-labeling algorithm in an EM framework. (c) Jointly train the feature model and the clustering head through the reliable pseudo-labeling algorithm.}
    \label{fig_framework}
\end{figure*}

\subsection{Unsupervised Representation Learning}
Unsupervised representation learning aims to map samples/images into semantically meaningful features without human annotations, which facilitates various down-stream tasks, such as object detection and classification. Previously, various pretext tasks were heuristically designed for this purpose, such as colorization~\cite{colorization2016}, rotation~\cite{rotation2018}, jigsaw~\cite{jigsaw2016}, etc.
Recently, contrastive learning methods combined with data augmentation strategies have achieved great success, such as SimCLR~\cite{simclr}, MOCO~\cite{He_2020_CVPR}, and BYOL~\cite{byol}, just to name a few.
On the other hand, clustering based representation learning methods have also achieved great progress.
Caron \etal~\cite{vf2018} proposed to alternatively performs k-means clustering algorithm on the entire dataset and train the classification network using the cluster labels.
Without explicitly computing cluster centers, Asano~\etal~\cite{Self-labelling} proposed a self-labeling approach that directly infers the pseudo-labels from the predicted cluster labels of the full dataset based on the Sinkhorn-Knopp algorithm~\cite{Sinkhorn}, and then uses the pseudo labels to the clustering network. 
Taking the advantages of contrastive learning, SwAV~\cite{caron2020swav} was proposed to simultaneously cluster the data while enforcing different transformations of the same image having the same cluster assignment.
It is worth emphasizing that, different from the unsupervised deep clustering methods in Section~\ref{sec_udc},  these clustering based representation learning methods aim to learn the representation features by clustering a much larger number of clusters than the number of ground-truth classes. This is consistent with our observation that directly clustering the target number of classes without accurate supervision will harm the representation learning due to the over-compression of instance-level features. Usually, to evaluate the quality of learned features, a linear classifier is independently trained with ground truth labels by freezing the parameters of representation learning models.

In this work, we aim to achieve unsupervised clustering with the exact number of real classes. On the other hand, we not only train the feature model but also the clustering head without using any annotations.
Actually, any unsupervised representation learning methods can be implemented as our feature model, which can be further improved via the joint training in SPICE.

\subsection{Semi-Supervised Classification}
Our method is also related to the semi-supervised classification methods as actually we reformulate the unsupervised clustering task into a semi-supervised learning paradigm in the joint training stage.
Semi-supervised classification methods aim to reduce the requirement of labeled data for training a classification model by providing a means of leveraging unlabeled data. In this category, remarkable results were obtained with consistency regularization~\cite{NIPS2016_30ef30b6, DBLP} that constrains the model to output the same prediction for different transformations of the same image, pseudo-labeling~\cite{pseudo} that uses confident predictions of the model as the labels to guide training processes, and entropy minimization~\cite{NIPS2004_96f2b50b, pseudo} that steers the model to output high-confidence predictions. MixMatch~\cite{mixmatch} algorithm combines these principles in a unified scheme and achieves an excellent performance, which is further improved by ReMixMatch~\cite{remixmatch} along this direction. Recently, FixMatch~\cite{fixmatch} proposed a simplified framework that uses the confident prediction of a weakly transformed image as the pseudo label when the model is fed a strong transformation of the same image, delivering superior results.

In this work, we target a more challenging task of training the clustering network without using any annotations, sometimes achieving comparable or even better results than the state-of-the-art semi-supervised learning methods.

\section{Method}
We aim to cluster a set of $N$ images $\mathfrak{X}=\{\vct{x}_i\}_{i=1}^N$ into $K$ classes by training a clustering network without using any annotations.
The clustering network can be conceptually divided into two parts: a feature model that maps images to feature vectors, $\vct{f}_i = \mathcal{F}(\vct{x}_i; \vct{\theta}_\mathcal{F})$, and a clustering head that maps feature vectors to the probabilities over $K$ classes, $\vct{p}_i = \mathcal{C}(\vct{f}_i; \vct{\theta}_\mathcal{C})$, where $\vct{\theta}_\mathcal{F}$ and $\vct{\theta}_\mathcal{C}$ represent the trainable parameters of the feature model $\mathcal{F}$ and the clustering head $\mathcal{C}$, respectively.
Different from the existing deep clustering methods, we use the outputs of  the feature model to measure the similarity among samples and use the clustering head to identify the discrepancy between clusters for pseudo-labeling, as shown in Fig.~\ref{fig_arcs}.
By effectively measuring both the similarity and discrepancy, we design two semantics-aware pseudo-labeling algorithms, prototype pseudo-labeling and reliable pseudo-labeling, to generate accurate and reliable self-supervision.

Specifically, we split the network training into three stages as shown in Fig.~\ref{fig_framework}.
First, we optimize the feature model $\mathcal{F}$ through the instance-level contrastive learning that enforces the features from different transformations of the same image being similar and the features from different images being discriminative from each other. 
Second, we optimize the clustering head $\mathcal{C}$ with the proposed prototype pseudo-labeling algorithm while freezing the feature model learned in the first stage.
Third, we optimize the feature model and the clustering head jointly with the proposed reliable pseudo-labeling algorithm.

In the following subsections, we introduce each training stage in detail.

\subsection{Feature Model Training with Contrastive Learning}\label{sec_fea}
To accurately measure the similarity of instance samples, here we adopt the instance discrimination based unsupervised representation learning method~\cite{He_2020_CVPR} for training the feature model.
As shown in Fig.~\ref{fig_framework}(a), there are two branches taking two random transformations of the same image as inputs, and each branch includes a feature model and a projection head that is a two-layer multilayer perceptron (MLP).
During training, we only optimize  the lower branch while the upper branch is updated as the moving average of the lower branch.
As contrastive learning methods benefit from a large training batch, a memory bank~\cite{wu2018unsupervised} is used to maintain a queue of encoded negative samples for reducing the requirement of GPU memory size, which is denoted as $\{\vct{z}^-_1, \vct{z}^-_2, \ldots, \vct{z}^-_{N_q} \}$, where $N_q$ is the queue size. 

Formally, given two transformations $\vct{x}'$ and $\vct{x}''$ of an image $\vct{x}$, the output of the upper branch is $\vct{z}^+ = \mathcal{P}(\mathcal{F}(\vct{x}'; \vct{\theta}'_\mathcal{F}); \vct{\theta}'_\mathcal{P})$, and the output of the lower branch is $\vct{z} = \mathcal{P}(\mathcal{F}(\vct{x}''; \vct{\theta}_\mathcal{F}); \vct{\theta}_\mathcal{P})$, where $\mathcal{P}$ denotes the projection head with the parameters $\vct{\theta}_\mathcal{P}$, and $\vct{\theta}'_\mathcal{F}$ and $\vct{\theta}'_{\mathcal{{P}}}$ are the moving averaging versions of $\vct{\theta}_\mathcal{F}$ and $\vct{\theta}_{\mathcal{{P}}}$.
The parameters $\vct{\theta}_\mathcal{{F}}$ and  $\vct{\theta}_\mathcal{P}$ are optimized with the following loss function:
\begin{equation}
    \label{eq_rl}
    \mathcal{L}_{fea} = -\log\left(\frac{\exp(\vct{z}\T\vct{z}^+/\tau)}{\sum_{i=1}^{N_q} \exp(\vct{z}\T\vct{z}^-_i/\tau)+\exp(\vct{z}\T\vct{z}^+/\tau)}\right),
\end{equation}
where the negative sample $\vct{z}^-_i$ may be computed from any images other than the current image $\vct{x}$, and $\tau$ is the temperature. Then, the parameters of the upper branch is updated as  $\vct{\theta}'_\mathcal{F} \leftarrow \mu \vct{\theta}'_\mathcal{F} + (1-\mu) \vct{\theta}_\mathcal{F}$, and $\vct{\theta}'_\mathcal{P} \leftarrow \mu \vct{\theta}'_\mathcal{P} + (1-\mu) \vct{\theta}_\mathcal{P}$, where $\mu\in[0,1)$ is a momentum coefficient.
The queue is updated by adding $\vct{z}^+$ to the end and removing the first item. All hyperparameters including $\tau=0.2$ and $\mu=0.999$ are the same as those in~\cite{He_2020_CVPR}.
The finally optimized feature model parameters are denoted as $\vct{\theta}_{\mathcal{F}}^s$, which will be used in the next stage.

\noindent\emph{Remark.} In practice, any unsupervised representation learning methods and network architectures can be applied in the SPICE framework.

\subsection{Clustering Head Training with Prototype Pseudo-Labeling}
\label{sec_proto}

Based on the trained feature model, here we train the clustering head by explicitly exploring both the similarity among samples and the discrepancy between clusters. 
Formally, given the image dataset $\mathfrak{X}$ and the feature model parameters $\vct{\theta}_\mathcal{F}^s$ obtained in Section~\ref{sec_fea}, we aim to train the clustering head only for predicting the cluster labels $\{y^s_i\}_{i=1}^N$.
The clustering head $\mathcal{C}$ is a two-layer MLP mapping the features to the probabilities, $\vct{p}_i = \mathcal{C}(\vct{f}_i; \vct{\theta}_\mathcal{C})$, where $\vct{f}_i = \mathcal{F}(\vct{x}_i; \vct{\theta}_\mathcal{F}^s)$.
However, in the unsupervised setting we do not have the ground truth for training.
To address this issue, we propose a prototype pseudo-labeling algorithm that alternatively estimates the pseudo labels of batch-wise samples and optimizes the parameters of the clustering head in an EM framework.

Generally, this training stage is to solve two sets of variables, i.e., the parameters of the clustering head $\mathcal{C}$, $\vct{\theta}_\mathcal{C}$, and the cluster labels $\{y^s_i\}$ of $\mathfrak{X}$ over $K$ clusters.
Analogous to k-means clustering algorithm~\cite{kmeans1967}, we solve two underlying sub-problems alternatively in an EM framework: the expectation (E) step is solving $\{y^s_i\}$ given $\vct{\theta}_\mathcal{C}$, and the maximization (M) step is solving $\vct{\theta}_\mathcal{C}$ given $\{y^s_i\}$.
Taking the advantages of contrastive learning, we clone the feature model into three branches as shown in Fig.~\ref{fig_framework}(b):
\begin{itemize}
\item The top branch takes original images as inputs and outputs the embedding features $\vct{f}_i$;
\item The middle branch takes the weakly transformed images as inputs and estimates the probabilities $\vct{p}_i$ over $K$ clusters, which is then combined with $\vct{f}_i$ to generate the pseudo labels $y^s_i$ through the proposed prototype pseudo-labeling algorithm; 
\item The bottom branch takes strongly transformed images as inputs and optimizes $\vct{\theta}_\mathcal{C}$ with the pseudo-labels. 
\end{itemize}

The EM process for the clustering head training is  detailed as follows. 

\noindent\textbf{Prototype Pseudo-Labeling (E-step).} The top branch computes the embedding features, $\mat{F} = [\vct{f}_1, \vct{f}_2,\ldots,\vct{f}_M]\T \in \mathbb{R}^{M\times D}$, of a mini-batch of samples $\mathfrak{X}_b$, and the middle branch computes the corresponding probabilities, $\mat{P}=[\vct{p}_1,\vct{p}_2,\ldots,\vct{p}_M]\T \in \mathbb{R}^{M\times K}$, for the weakly transformed samples, $\alpha(\mathfrak{X}_b)$; here, $M$ is mini-batch size, $D$ is the dimension of the feature vector, and $\alpha$ denotes the weak transformation over the input image.

Given $\mat{P}$ and $\mat{F}$, the top confident samples are selected to estimate the prototypes for each cluster, and then the indices of the cluster prototypes are assigned to their nearest neighbors as the pseudo labels.
Formally, the top confident samples for each cluster, taking the $k$-th cluster as an example, are selected as:
\begin{align}
    \mathfrak{F}_k = \{ \vct{f}_i | i\in\text{argtopk}(\mat{P}_{:,k}, \tfrac{M}{K}), \forall\ i=1, 2, \ldots, M\},\label{eq_conf}
\end{align}
where $\mat{P}_{:,k}$ denotes the $k$-th column of matrix $\mat{P}$ and $\text{argtopk}\left(\mat{P}_{:,k}, \tfrac{M}{K}\right)$ returns the top $\tfrac{M}{K}$ confident sample indices from $\mat{P}_{:,k}$.
Naturally, the cluster centers $\{\vct{\gamma}_k\}_{k=1}^K$ in the embedding space are computed as
\begin{align}
    \label{eq_center}
    \vct{\gamma}_k = \frac{K}{M} \sum_{\vct{f}_i \in \mathfrak{F}_k} \vct{f}_i, \quad\forall\ k=1, 2,\ldots, K.
\end{align}
By computing the cosine similarity between embedding features $\vct{f}_i$ and the cluster center $\vct{\gamma}_k$, we select $\frac{M}{K}$ nearest samples to $\vct{\gamma}_k$, denoted by $\mathfrak{X}^k$, to have the same cluster label, $y_i^s=k, \forall \vct{x}_i \in \mathfrak{X}^k$. Thus, a mini-batch of images with semantic pseudo-labels, $\mathfrak{X}^s$, is constructed as
\begin{equation}
    \label{eq_label}
    \mathfrak{X}^s = \{(\vct{x}_i, y_i^s) | \forall \vct{x}_i \in \mathfrak{X}^k,  k =1,2,\ldots, K\}.
\end{equation}

\begin{figure}[bt!]
    \centering
    \includegraphics[width=0.48\textwidth]{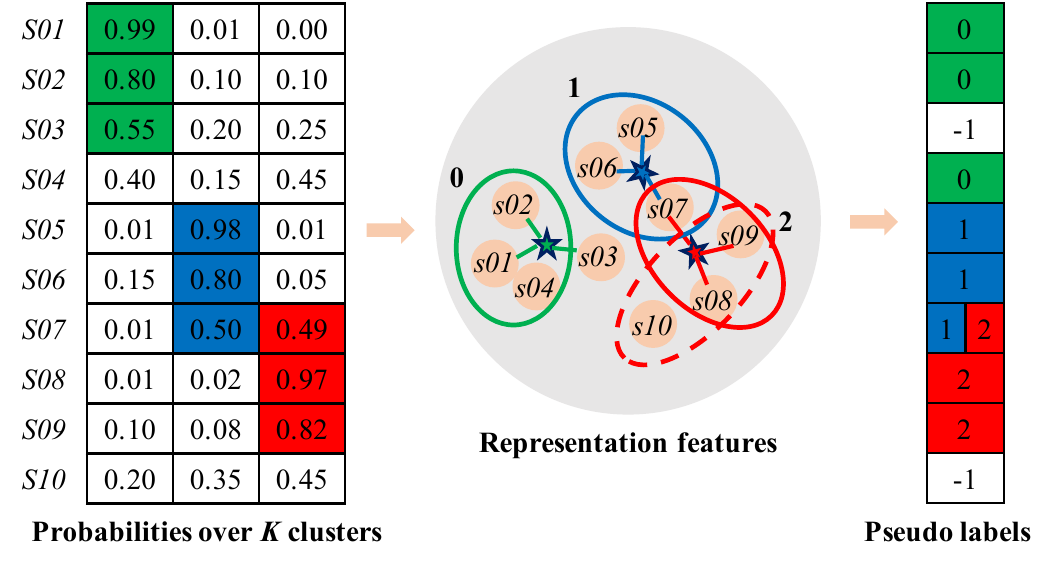}
    \caption{A toy example for prototype pseudo-labeling. First, given the predicted probabilities of 10 samples over 3 clusters, top 3 confident samples are selected for each cluster, marked as green, blue, and red colors respectively. Then, the selected samples are mapped into the corresponding features (denoted by dots) to estimating the prototypes (denoted by stars) for each cluster, where stars are estimated with connected dots. Finally, the top 3 nearest samples to each cluster prototype (the dots within the same ellipse) are selected and assigned with the index of the corresponding prototype. Other unselected samples are signed with -1 and will not be used for training. The dashed ellipse denotes non-overlap assignment.}
    \label{fig_sl}
\end{figure}

A toy example of the prototype pseudo-labeling process is shown in Fig.~\ref{fig_sl}, where there is a batch of 10 samples and 3 clusters, 3 confident samples for each cluster are selected according to the predicted probabilities to calculate the prototypes in the feature space, and 3 nearest samples to each cluster are selected and labeled. Note that there may exist overlapped samples between different clusters, so there are two options to handle these labels: one is the overlap assignment that one sample may have more than one cluster labels as indicated by the blue and red circles in Fig.~\ref{fig_sl}, and the other is non-overlap assignment that all samples have only one cluster label as indicated by the dished red circle. We found that the overlap assignment is better as analyzed in Section~\ref{sec_abl}.

\noindent\textbf{Training Clustering Head (M-step).} 
Given the labeled samples $\mathfrak{X}^s$, the clustering head parameters are optimized  in a supervised learning manner.
Specifically, 
we compute the probabilities of strong transformations $\beta(\mathfrak{X}^s)$  in the forward pass, where $\beta$ denotes the strong augmentation operator.
Then, the clustering head $\mathcal{C}$ can be optimized in the backward pass by minimizing the following cross-entropy (CE) loss:
\begin{align}
    \label{eq_loss}
    \mathcal{L}_{clu} = \frac{1}{M} \sum_{i=1}^{M} \mathcal{L}_{ce}(y_i^s, \vct{p}'_{i}), 
\end{align}
where $\vct{p}'_{i} = \text{softmax}(\vct{p}_{i})$ and  
        $\vct{p}_{i} =  \mathcal{A}(\mathcal{F}(\beta(\vct{x}_i); \vct{\theta}_\mathcal{F}^s); \vct{\theta}_\mathcal{A})$, 
$\mathcal{L}_{ce}$ is the cross-entropy loss function.

In Eq.~\eqref{eq_loss}, we use double softmax functions before computing CE loss, as $\vct{p}_i$ is already the outputs of a softmax function.
Considering that the pseudo-labels are not as accurate as the ground truth labels, the basic idea behind the double softmax implementation is to reduce the learning speed especially when the predictions are of low probabilities, which can benefit the dynamically clustering process during training 
(please see Appendix~A for the detailed analysis of the double softmax implementation).

The above process for training the clustering head is summarized in Algorithm~\ref{alg_1}.

\begin{algorithm}
\scriptsize
\label{alg_1}
\caption{Training Clustering Head.}
\LinesNumbered
\KwIn{Dataset $\mathfrak{X}=\{\vct{x}_i\}_{i=1}^N$, $\vct{\theta}_\mathcal{F}^s$, $K$, $M$, $m$, $T$, $\alpha$, $\beta$}
\KwOut{Cluster label $y_i^s$ of $x_i \in \mathcal{X}$}
Set feature model parameters to $\vct{\theta}_\mathcal{F}^s$, $t = 0$, and initialize $\vct{\theta}_\mathcal{C}$ \;
\While{$t < $ T}{

    \For{$b = 1, 2, \dots, \lfloor\frac{N}{M}\rfloor$}{
            \textbf{\emph{E-step}:} \\
            Select $M$ samples  from $\mathfrak{X}$ as $\mathfrak{X}_b$\;
            Compute embedding features $\mat{F} = \mathcal{F}(\mathcal{X}_b; \vct{\theta}_\mathcal{F}^s)$ \;
            Predict probabilities $\mat{P} = \mathcal{C}(\mathcal{F}(\alpha(\mathfrak{X}_b); \vct{\theta}_\mathcal{F}^s); \vct{\theta}_\mathcal{C})$ \;
            Construct labeled image set $\mathcal{X}^s$ with Eqs.~\eqref{eq_conf},~\eqref{eq_center}, and~\eqref{eq_label} \;
            \textbf{\emph{M-step}:} \\

            Compute probabilities $\mat{P} = \mathcal{C}(\mathcal{F}(\beta(\mathcal{X}_b); \vct{\theta}_\mathcal{F}^{s}); \vct{\theta}_\mathcal{C})$ \;
            Optimize $\vct{\theta}_\mathcal{C}$ by minimizing Eq.~\eqref{eq_loss} \;

    }
    $t \leftarrow t + 1$
}

Select the best clustering head with the minimum loss as $\vct{\theta}^s_\mathcal{C}$ \;

\ForEach{$\vct{x}_i \in \mathfrak{X}$}{
        $\vct{p}_i = \mathcal{C}(\mathcal{F}(\vct{x}_i; \vct{\theta}_\mathcal{F}^{s}); \vct{\theta}^s_\mathcal{C}])$ \;
        $y_i^s=\argmax_k(\vct{p}_i)$\;
}
\end{algorithm}

\noindent\emph{Remark.} In this stage, we fix the parameters of feature models from the representation learning, and only optimize the light-weight clustering head. Thus, the computational burden is significantly reduced so that we can train multiple clustering heads simultaneously and independently. By doing so, the instability of clustering from the initialization can be effectively alleviated through selecting the best 
clustering head. Specifically, the best head with the parameters $\vct{\theta}_\mathcal{C}^s$ can be selected for the minimum loss value of $\mathcal{L}_{clu}$ over the whole dataset; \ie, we set $M=N$ and follow E-step and M-step in Algorithm~\ref{alg_1} to compute the loss value.
During testing, the best clustering head is used to cluster the input images into different clusters.

\subsection{Joint Training with Reliable Pseudo-Labeling}
\label{sec_semi}

The feature model and the clustering head are optimized separately so far, which tends to be a sub-optimal solution.
On the one hand, the imperfect feature model may lead to some similar features corresponding to really different clusters; thus, assigning neighbor samples with the same pseudo-label is not always reliable.
On the other hand, the imperfect clustering head may assign really dissimilar samples with the same cluster label, such that only using the predicted labels for fine-tuning is also not always reliable.
To overcome these problems, we design a reliable pseudo-labeling algorithm to train the feature model and clustering head  jointly for further improving the clustering performance.

\noindent\textbf{Reliable Pseudo-Labeling.}
Given the embedding features and the predicted labels $\{(\vct{x}_i, \vct{f}_i, y_i^s)\}_{i=1}^N$ obtained in Section~\ref{sec_proto}, we select $N_s$ nearest samples for each sample $\vct{x}_i$ according to the cosine similarity between embedding features. The corresponding labels of these nearest samples are denoted by $\mathfrak{L}_{i}$. Then, the semantically consistent ratio $r_i$ of the sample $\vct{x}_i$ is defined as
\begin{align}
\label{eq_con}
r_i = \frac{1}{N_s} \sum_{y \in \mathfrak{L}_{i}} \mathbbm{1}(y = y^s_i).
\end{align}
Given a predefined threshold $\lambda$, if $r_i > \lambda$, the sample $(\vct{x}_i, y_i^{s})$ is identified as the reliably labeled for joint training, and otherwise the corresponding label is ignored.
Through the reliable pseudo-labeling, a subset of reliable samples $\mathfrak{X}^r$ are selected as: 
\begin{align}
\mathfrak{X}^r = \{(\vct{x}_i, y_i^s) | r_i > \lambda, \  \forall\ i=1, 2,\ldots, N\}.
\end{align}

\noindent\textbf{Joint Training.}
Given the above partially labeled samples, the clustering problem can be converted into a semi-supervised learning paradigm to train the clustering network jointly.
Here we adapt a simple semi-supervised learning method~\cite{fixmatch}.
During training, the subset of reliably labeled samples keep fixed.
On the other hand, all training samples should be consistently clustered, \ie, different transformations of the same image are constrained to have the consistent prediction.
To this end, as shown in Fig.~\ref{fig_framework}(c), the confidently predicted label of weak transformations is used as the pseudo-label for strong transformations of the same image.
Formally, the consistency pseudo label $y_j^u$ of the sample $\vct{x}_j$ is calculated as in Eq.~\eqref{eq_cl}:
\begin{equation}
\label{eq_cl}
y^u_j = 
\begin{cases}
  \argmax(\vct{p}_j) & \text{if} \max(\vct{p}_j) \ge \eta, \\
  -1 & \text{otherwise}
\end{cases}
\end{equation}
where $\vct{p}_j = \mathcal{C}(\mathcal{F}(\alpha(\vct{x}_j); \vct{\theta}_\mathcal{F}); \vct{\theta}_\mathcal{C}))$, and $\eta$ is the confidence threshold.

Then, the whole network parameters $\vct{\theta}_\mathcal{F}$ and $\vct{\theta}_\mathcal{C}$ are optimized with the following loss function:

\begin{align}
\mathcal{L}_{joint}& = \frac{1}{L} \sum_{i=1}^L \underbrace{\mathcal{L}_{ce}(y_i^s, \mathcal{C}(\mathcal{F}(\alpha(\vct{x}_i); \vct{\theta}_\mathcal{F}); \vct{\theta}_\mathcal{C}))}_{\text{partial samples with reliable pseudo-labels}}  \label{eq_loss_semi}\\
                    &+\frac{1}{U} \sum_{j=1}^{U} \underbrace{\mathbbm{1}(y_j^u \ge 0) \mathcal{L}_{ce}(y_j^u, \mathcal{C}(\mathcal{F}(\beta(\vct{x}_j); \vct{\theta}_\mathcal{F}); \vct{\theta}_\mathcal{C}))}_{\text{all samples with consistency pseudo-labels}},\notag
\end{align}
where the first term is computed with reliably labeled samples $(\vct{x}_i, y_i^s)$ drawn from $\mathfrak{X}^r$, and the second term is computed with pseudo-labeled samples $(\vct{x}_i, y_i^u)$ drawn from the whole dataset $\mathfrak{X}$, which is dynamically labeled by thresholding the confident predictions as in Eq.~\eqref{eq_cl}.
$L$ and $U$ denote the numbers of labeled and unlabeled images in a mini-batch.

\noindent\emph{Remark.} 
Although we adapt the FixMatch~\cite{fixmatch} semi-supervised classification method for  image clustering in this work, we highlight that other semi-supervised algorithms can also be used here with reliable samples generated by the proposed reliable pseudo-labeling algorithm.


\emph{Note that SPICE sheds light on the importance of utilizing both instance-level similarity and semantic level discrepancy for clustering.
With this key idea in mind, this study focuses on developing the semantic pseudo-labeling algorithms that can actually estimate the instance similarity and semantic discrepancy for better clustering results.
Actually, SPICE presents a general unsupervised clustering framework that gradually trains the feature model, clustering head, and whole model end-to-end.
This framework is able to organically unify advanced unsupervised representation learning and semi-supervised learning methods for clustering through the proposed semantic pseudo-labeling method.}

\section{Experiments and Results}

\subsection{Benchmark Datasets and Evaluation Metrics}

\begin{table}[ht]
\footnotesize
\renewcommand{\arraystretch}{1.3}
\caption{Specifications and partitions of selected datasets.}
\label{table_data}
\centering
\resizebox{\linewidth}{!}{
\begin{tabular}{lcccc}
\toprule
Dataset & Image size & \# Training & \# Testing & \# Classes (K) \\
\midrule
STL10         & $96\times96$   & 5,000      & 8,000   &  10   \\
CIFAR-10       & $32  \times 32$   & 50,000     & 10,000  &  10   \\
CIFAR-100-20   & $32  \times 32$   & 50,000     & 10,000  &  20   \\
ImageNet-10   & $224 \times 224$  & 13,000     & N/A    &  10   \\
ImageNet-Dog  & $224 \times 224$  & 19,500     & N/A    &  15   \\
Tiny-ImageNet & $64  \times 64$   & 100,000    & 10,000  &  200  \\
\bottomrule

\end{tabular}}
\end{table}

We evaluated the performance of SPICE on six commonly used image clustering datasets, including
STL10, CIFAR-10, CIFAR-100-20, ImageNet-10, ImageNet-Dog, and Tiny-ImageNet.
The key details of each dataset are summarized in Table~\ref{table_data}, where the datasets reflect a diversity of image sizes, the number of images, and the number of clusters.
Different existing methods used different image sizes for training and testing; for example, CC~\cite{cc} resizes all images of these six datasets into $224 \times 224$, and GATCluster~\cite{gatcluster} studies the effectiveness of different image sizes on ImageNet-10 and ImageNet-Dog, showing that too large or too small may harm the clustering performance.
In this work, we naturally use the original size of images without resizing to a larger size of images. For the ImageNet, we adopt the commonly used image size of $224 \times 224$.

Three popular metrics are used to evaluate clustering results, including Adjusted Rand Index (ARI)~\cite{hubert1985comparing}, Normalized Mutual Information (NMI)~\cite{strehl2002clusterensembles}, and clustering Accuracy (ACC)~\cite{LiD06}.

\begin{table*}[t]
\footnotesize
\renewcommand{\arraystretch}{1.3}
\renewcommand\tabcolsep{3.6pt}
\caption{Comparison with competing methods that were trained and tested on the whole dataset  (train and test split datasets are merged as one). Note that SPICE was trained on the target datasets only without using extra data and without using any annotations, which are the exactly the same as data used in the existing methods. The best results are highlighted in \textbf{bold}.}
\label{table_results_cluster}
\centering
\resizebox{\textwidth}{!}{
\begin{tabular}{ rcccccccccccccccccc}
\toprule

\multirow{2}{*}{Method}             & \multicolumn{3}{c}{STL10}&\multicolumn{3}{c}{ImageNet-10}&\multicolumn{3}{c}{ImageNet-Dog-15}&\multicolumn{3}{c}{CIFAR-10}&\multicolumn{3}{c}{CIFAR-100-20}&\multicolumn{3}{c}{Tiny-ImageNet-200}\\
                                                                        \cline{2-19}
                                                                        & ACC  & NMI  & ARI& ACC&NMI&ARI   & ACC&NMI&ARI  &ACC&NMI&ARI   &ACC&NMI&ARI  &ACC&NMI&ARI\\
\midrule
k-means~\cite{kmeans1967}                & 0.192  & 0.125 & 0.061   & 0.241 &0.119& 0.057   & 0.105 &0.055&0.020   & 0.229 &0.087&0.049   & 0.130 &0.084&0.028   & 0.025 &0.065&0.005\\
SC~\cite{Spectral2002}                   & 0.159  & 0.098 & 0.048   & 0.274 &0.151&0.076    & 0.111 &0.038&0.013   & 0.247 &0.103&0.085   & 0.136 &0.090&0.022   & 0.022 &0.063&0.004\\
AC~\cite{Pasi2006Fast}                   & 0.332  & 0.239 & 0.140   & 0.242 &0.138&0.067    & 0.139 &0.037&0.021   & 0.228 &0.105&0.065   & 0.138 &0.098&0.034   & 0.027 &0.069&0.005\\
NMF~\cite{NMF}                           & 0.180  & 0.096 & 0.046   & 0.230 &0.132&0.065    & 0.118 &0.044&0.016   & 0.190 &0.081&0.034   & 0.118 &0.079&0.026   & 0.029 &0.072&0.005\\
AE~\cite{Bengio2007Greedy}               & 0.303  & 0.250 & 0.161   & 0.317 &0.210&0.152    & 0.185 &0.104&0.073   & 0.314 &0.239&0.169   & 0.165 &0.100&0.048   & 0.041 &0.131&0.007\\
SDAE~\cite{SDAE2010}                     & 0.302  & 0.224 & 0.152   & 0.304 &0.206&0.138    & 0.190 &0.104&0.078   & 0.297 &0.251&0.163   & 0.151 &0.111&0.046   & 0.039 &0.127&0.007\\
DCGAN~\cite{dcgan}                       & 0.298  & 0.210 & 0.139   & 0.346 &0.225&0.157    & 0.174 &0.121&0.078   & 0.315 &0.265&0.176   & 0.151 &0.120&0.045   & 0.041 &0.135&0.007\\
DeCNN~\cite{Zeiler2010Deconvolutional}   & 0.299  & 0.227 & 0.162   & 0.313 &0.186&0.142    & 0.175 &0.098&0.073   & 0.282 &0.240&0.174   & 0.133 &0.092&0.038   & 0.035 &0.111&0.006\\
VAE~\cite{vae}                           & 0.282  & 0.200 & 0.146   & 0.334 &0.193&0.168    & 0.179 &0.107&0.079   & 0.291 &0.245&0.167   & 0.152 &0.108&0.040   & 0.036 &0.113&0.006\\
JULE~\cite{Yang2016Joint}                & 0.277  & 0.182 & 0.164   & 0.300 &0.175&0.138    & 0.138 &0.054&0.028   & 0.272 &0.192&0.138   & 0.137 &0.103&0.033   & 0.033 &0.102&0.006\\
DEC~\cite{Xie2016}                       & 0.359  & 0.276 & 0.186   & 0.381 &0.282&0.203    & 0.195 &0.122&0.079   & 0.301 &0.257&0.161   & 0.185 &0.136&0.050   & 0.037 &0.115&0.007\\
DAC~\cite{DAIC2017}                      & 0.470  & 0.366 & 0.257   & 0.527 &0.394&0.302    & 0.275 &0.219&0.111   & 0.522 &0.396&0.306   & 0.238 &0.185&0.088   & 0.066 &0.190&0.017\\
DeepCluster~\cite{vf2018}                & 0.334  & N/A   &  N/A    & N/A &N/A&N/A          & N/A &N/A&N/A         & 0.374& N/A   &  N/A  & 0.189 & N/A &  N/A   & N/A &N/A&N/A\\
DDC~\cite{DDC2019}                       & 0.489  & 0.371 & 0.267   & 0.577 &0.433&0.345    & N/A &N/A&N/A         & 0.524 &0.424&0.329   & N/A &N/A&N/A         & N/A &N/A&N/A\\
IIC ~\cite{IIC2019}                      & 0.610  & N/A   &  N/A    & N/A &N/A&N/A          & N/A &N/A&N/A         & 0.617& N/A & N/A     & 0.257& N/A & N/A     & N/A &N/A&N/A\\
DCCM~\cite{Wu_2019_ICCV}                 & 0.482  & 0.376 & 0.262   & 0.710 &0.608 &0.555   & 0.383&0.321 &0.182   & 0.623& 0.496&0.408   & 0.327 &0.285&0.173   & 0.108 &0.224&0.038\\
DSEC~\cite{dsec}                         & 0.482  & 0.403 & 0.286   & 0.674 &0.583&0.522    & 0.264 &0.236&0.124   & 0.478 &0.438&0.340   & 0.255 &0.212&0.110   & 0.066 &0.190&0.017\\
GATCluster~\cite{gatcluster}             & 0.583  & 0.446 & 0.363   & 0.762 &0.609 &0.572   & 0.333&0.322 & 0.200  & 0.610&0.475 &0.402   & 0.281 &0.215&0.116   & N/A &N/A&N/A\\
PICA~\cite{Huang_2020_CVPR}              & 0.713  & 0.611 & 0.531   & 0.870 &0.802 &0.761   & 0.352&0.352 & 0.201  & 0.696&0.591 &0.512   & 0.337 &0.310&0.171   & 0.098 &0.277&0.040\\
CC~\cite{cc}                             & 0.850  & 0.746 & 0.726   & 0.893 &0.859 &0.822   & 0.429&0.445 & 0.274  & 0.790&0.705 &0.637   & 0.429 &0.431&0.266   & 0.140 &0.340&0.071\\
IDFD~\cite{idfd}                         & 0.756  & 0.643  & 0.575  & 0.954 &0.898 &0.901   & 0.591&0.546 & 0.413  & 0.815& 0.711 &0.663  & 0.425 &0.426&0.264   & N/A &N/A&N/A\\
\hline
\textbf{SPICE$_{\mathrm{s}}$}                  & 0.908  & 0.817 & 0.812   & 0.921 &0.828 &0.836 & 0.646&0.572&0.479   & 0.838&0.734  &0.705    & 0.468 &0.448&0.294      & \textbf{0.305} &\textbf{0.449}&\textbf{0.161}\\
\textbf{SPICE}                        & \textbf{0.938}&\textbf{0.872}&\textbf{0.870} & \textbf{0.959} &\textbf{0.902} &\textbf{0.912} & \textbf{0.675} & \textbf{0.627} & \textbf{0.526}    & \textbf{0.926}&\textbf{0.865}&\textbf{0.852}   &\textbf{0.538}& \textbf{0.567}&\textbf{0.387}     & N/A &N/A&N/A\\
\bottomrule

\end{tabular}
}
\end{table*}

\subsection{Implementation Details}
For fair comparison, we mainly adopted two backbone networks, \ie ResNet18 and ResNet34~\cite{ResNet2015}, for representation learning. 
The classification head in SPICE consists of two fully-connected layers; \ie, D-D-K, where $D$ and $K$ are the dimension of features and the number of clusters, respectively.
Specifically, $D=512$ for both ResNet18 and ResNet34 backbone networks, and the cluster number $K$ is predefined as the number of classes on the target dataset as shown in Table~\ref{table_data}.
To show how the joint training (third stage) improves the clustering performance, we refer to SPICE without joint training as SPICE$_{\mathrm{s}}$, where
the subscript $\mathrm{s}$ indicates the separate training.

For representation learning, we use MoCo-v2~\cite{He_2020_CVPR} in all our experiments, which was also used in SCAN~\cite{scan}.
For weak augmentation, a standard flip-and-shift augmentation strategy is implemented as in FixMatch~\cite{fixmatch}.
For strong augmentation, we adopt the same strategies used in SCAN~\cite{scan}. Specifically, the images were strongly augmented by composing Cutout~\cite{cutout} and four randomly selected transformations from RandAugment~\cite{randaugent}.

In SPICE, we use 10 clustering heads, and select the best head with the minimum loss for the final head in each trial.
To select the reliably labeled images in SPICE through reliable pseudo-labeling, we empirically set $N_s=100$ and $\lambda = 0.95$.
The hyperparameters involved in FixMatch~\cite{fixmatch} keep the same as those in the original paper, including $\eta = 0.95$.
We set the batch size $M$ to 1,000.


\subsection{Clustering Performance Comparison}
\label{sec_resutls_clustering}
The existing methods can be divided into two groups according to their training and testing settings.
One is to train and test the clustering model on the whole dataset combining the train and test splits as one.
The other is to train and test the clustering model on the separate train and test datasets.
For fair comparison, the proposed SPICE is evaluated under both two settings and compared with the existing methods accordingly.

Table~\ref{table_results_cluster} shows the comparison results of clustering on the whole dataset.
In reference to the recently developed methods~\cite{IIC2019, gatcluster, cc}, the same backbone, \ie ResNet34, was used during learning the feature model and clustering head. Duplicating the original settings in FixMatch~\cite{fixmatch}, we used WideResNet-28-2 for CIFAR-10, WideResNet-28-8 for CIFAR-100-20, and WideResNet-37-2 for STL-10. For ImageNet-10, ImageNet-Dog, and Tiny-ImageNet datasets that were not used in FixMatch, we simply used the same ResNet34 during joint learning.
The results show that SPICE improves ACC, NMI, and ARI by 8.8\%, 12.6\%, and 14.4\% respectively over the previous best results that were recently reported by CC~\cite{cc} on STL10. On average, our proposed method also improves ACC, NMI, and ARI by about 10\% on ImageNet-Dog-15, CIFAR-10, CIFAR-100-20, and Tiny-ImageNet-200.
It is worth emphasizing that, without the joint training stage, SPICE$_\mathrm{s}$ still performs better than the exiting deep clustering methods using the same network architecture on the most of the datasets.
\emph{These results convincingly show the superior performance of the proposed method using exactly the same backbone networks and datasets.}
The final joint training results are obviously better than those from the separate training on all datasets, especially on CIFAR-10 (improved by 8.8\% for ACC) and CIFAR-100-20 (improved by 7.0\% for ACC).
In clustering the images in Tiny-ImageNet-200, although our results are significantly better than the existing results, while still very low. This is mainly due to the class hierarchies; \ie, some classes share the same supper class, as analyzed in~\cite{scan}. Due to the low performance, some clusters cannot be reliably labeled based on the reliable pseudo-labeling algorithm so that end-to-end training cannot be applied for further boosting clustering performance. Thus, it is still an open problem for clustering a large number of hierarchical clusters.


\begin{table}[t]
\footnotesize
\renewcommand{\arraystretch}{1.3}
\renewcommand\tabcolsep{1.5pt}
\caption{Comparison with competing methods on split datasets (training and testing images are mutually exclusive). For fair comparison, both SCAN$_\text{MoCo}$ and SPICE used MoCo for feature learning, and ResNet18 as backbone in all training stages. Here the best results for all methods were used for comparison. SPICE$_\text{Res34}$ used the ResNet34 as backbone. The best two unsupervised results are highlighted in \textbf{bold}.}
\label{table_results_cls}
\centering
\resizebox{\linewidth}{!}{
\begin{tabular}{lccccccccc}
\toprule

\multirow{2}{*}{Method}             & \multicolumn{3}{c}{STL10}&\multicolumn{3}{c}{CIFAR-10}&\multicolumn{3}{c}{CIFAR-100-20}\\
                                                                        \cline{2-10}
                                    & ACC  & NMI  & ARI   &ACC&NMI&ARI   &ACC&NMI&ARI \\
\midrule
ADC~\cite{ADC2019}                & 0.530 & N/A &N/A            & 0.325 & N/A& N/A     &0.160  &N/A&N/A        \\
TSUC~\cite{tsuc}                  & 0.665  & N/A & N/A         & 0.810&N/A &N/A       & 0.353 &N/A&N/A  \\
NNM~\cite{Dang_2021_CVPR}                   & 0.808  & 0.694 & 0.650      & 0.843&0.748 &0.709   & 0.477 &0.484&0.316  \\
SCAN~\cite{scan}                  & 0.809  & 0.698 & 0.646      & 0.883&0.797 &0.772   & 0.507 &0.486&0.333  \\
RUC$_\text{SCAN}$~\cite{Park_2021_CVPR}              & 0.867  & N/A & N/A          & 0.903&N/A &N/A       & 0.533 &N/A&N/A  \\

SCAN$_\text{MoCo}$~\cite{scan}                           & 0.855  & 0.758 & 0.721      & 0.874&0.786 &0.756   & 0.455 &0.472&0.310   \\

\hline
SPICE$_\mathrm{s}$               & 0.862  & 0.756 & 0.732          & 0.845&0.739 &0.709   & 0.468 & 0.457 & 0.321 \\
\textbf{SPICE}                    & \textbf{0.920}  & \textbf{0.852} & \textbf{0.836}         & \textbf{0.918} & \textbf{{}0.850} & \textbf{0.836}  & \textbf{0.535} & \textbf{0.565} & \textbf{0.404} \\

\textbf{SPICE$_\text{Res34}$}                   & \textbf{0.929}  & \textbf{0.860} & \textbf{0.853}   & \textbf{0.917} & \textbf{0.858} & \textbf{0.836}  & \textbf{0.584} & \textbf{0.583} & \textbf{0.422}      \\

\hline

Supervised                        & 0.806 & 0.659 & 0.631        & 0.938 & 0.862 & 0.870  & 0.800 & 0.680 & 0.632 \\
\bottomrule

\end{tabular}}
\end{table}
Table~\ref{table_results_cls} shows the comparison results of clustering on the split train and test datasets.
For fair comparison, we re-implement SCAN with MoCo~\cite{He_2020_CVPR} for representation learning, denoted as SCAN$_\text{MoCo}$.
It can be seen that the results of SCAN$_\text{MoCo}$ on SLT10 are obviously better than SCAN, while the performance on CIFAR-10 and CIFAR-100-20 drops slightly.
Compared with the baseline method SCAN$_\text{MoCo}$, SPICE improves ACC, NMI, and ARI by 6.5\%, 9.4\%, and 11.5\% on STL10, by 4.4\%, 6.4\%, and 8.0\% on CIFAR-10, and by 8.0\%, 9.3\%, and 9.4\% on CIFAR-100-20, under the exactly same setting.
Without joint learning, SPICE$_\mathrm{s}$ already performs better than SCAN$_\text{MoCo}$ that contains the pretraining, clustering, and finetuning stages on STL10 and CIFAR100-20.
Moreover, we evaluate the SPICE using larger backbone networks as used in the whole dataset setting, the results on STL10 and CIFAR-10 are very similar while the results on CIFAR-100-20 are significantly improved.
Remarkably, SPICE significantly reduces the gap between unsupervised and supervised classification. On the STL10 that includes both labeled and unlabeled images, the results of unsupervised methods are better than the supervised as the unsupervised methods can leverage the unlabeled images for representation learning while the supervised cannot.
On CIFAR-10 and CIFAR-100-20, all methods using the same images for training and the proposed SPICE further reduces the performance gap compared with the supervised counterpart, particularly, only 2\% for ACC gap on CIFAR-10.

\begin{table}[ht]
\footnotesize
\renewcommand{\arraystretch}{1.2}
\renewcommand\tabcolsep{6.8pt}
\caption{More detailed comparison results on STL10. Here all methods were trained and tested on the split train and test datasets respectively. Both the mean and standard deviation results were reported. Each method was conducted five times. Here all methods used the ResNet18 backbone, SCAN$_\text{MoCo}$ and SPICE used MoCo for feature learning with STL10 images only. SCAN$_\text{MoCo}^*$ means no self-labeling.}
\label{table_results_detail}
\centering
\resizebox{\linewidth}{!}{
\begin{tabular}{lccc}

\toprule

Method                          & ACC&NMI&ARI\\

\midrule
Supervised                      & 0.806 & 0.659 & 0.631    \\

MoCo+k-means                  & 0.797$\pm$0.046 & 0.768$\pm$0.021 & 0.624$\pm$0.041     \\

SCAN$_\text{MoCo}^*$                        & 0.787$\pm$0.036 & 0.697$\pm$0.026 & 0.639$\pm$0.041    \\

SCAN$_\text{MoCo}$                        & 0.797$\pm$0.034 & 0.701$\pm$0.032 & 0.649$\pm$0.044    \\

SPICE$_\mathrm{s}$                      & 0.852$\pm$0.011 & 0.749$\pm$0.008 & 0.719$\pm$0.015    \\

SPICE                           & 0.918$\pm$0.002 & 0.849$\pm$0.003 & 0.836$\pm$0.002   \\
\bottomrule
\end{tabular}}
\end{table}
Table~\ref{table_results_detail} provides more detailed comparison results on STL10, where all these three methods use MoCo~\cite{He_2020_CVPR} for representation learning and were conducted five times for computing the mean and standard deviation of the results.
Compared with SCAN$_\text{MoCo}^*$ that explores the instance similarity only without fine-tuning, SPICE$_\mathrm{s}$ explicitly leverages both the instance similarity and semantic discrepancy for learning clusters. On the other hand, different from k-means that infers the cluster labels with cluster centers, SPICE$_\mathrm{s}$ uses the non-linear clustering head to predict the cluster labels.
It can be seen that SPICE$_\mathrm{s}$ is significantly better than MoCo+k-means and SCAN$_\text{MoCo}^*$  in terms of both the mean and standard deviation metrics, demonstrating the superiority of the proposed prototype pseudo-labeling algorithm.
Also, the final results of SPICE is significantly better than those of SCAN$_\text{MoCo}$ in terms of mean performance and the stability.

Overall, our comparative results systematically demonstrate the superiority of the proposed SPICE method on both the whole and split dataset settings.

\subsection{Semi-Supervised Classification Comparison}

\begin{table*}[ht]
\renewcommand{\arraystretch}{1.3}
\caption{Comparison results with semi-supervised learning on STL10 and CIFAR-10. The best three results are in \textbf{bold}.}
\label{table_results_semi}
\centering
\resizebox{\textwidth}{!}{
\begin{tabular}{ccccccccccc}

\toprule
\multirow{3}{*}{Method}             & \multicolumn{7}{c}{Semi-supervised}&&\multicolumn{2}{c}{Unsupervised}\\
                                                                        \cline{2-8}\cline{10-11}
                          & \tabincell{c}{$\Pi$-Model\\ \cite{ladder}}& \tabincell{c}{Pseudo-Labeling\\\cite{pseudo-label}}& \tabincell{c}{Mean Teacher\\\cite{mean-teacher}}& \tabincell{c}{MixMatch\\\cite{mixmatch}}&\tabincell{c}{UDA\\\cite{UDA}}&\tabincell{c}{ReMixMatch\\\cite{remixmatch}}&\tabincell{c}{FixMatch\\\cite{fixmatch}}&& \tabincell{c}{SCAN\\\cite{scan}} & \tabincell{c}{SPICE\\(ours)}\\

\midrule
STL10                           & 0.748$\pm$0.008 & 0.720$\pm$0.008 & 0.786$\pm$0.024 & 0.896$\pm$0.006 & \textbf{0.923$\pm$0.005} & \textbf{0.948$\pm$0.005}  & 0.920$\pm$0.015 && 0.767$\pm$0.019 & \textbf{0.929$\pm$0.001}   \\

CIFAR-10                         & 0.457$\pm$0.040 & 0.502$\pm$0.004 & 0.677$\pm$0.009 & 0.890$\pm$0.009  & 0.912$\pm$0.011 & \textbf{0.947$\pm$0.001} & \textbf{0.949$\pm$0.003} && 0.876$\pm$0.004 & \textbf{0.917$\pm$0.002}   \\
\bottomrule

\end{tabular}}
\end{table*}

In this subsection, we further compare SPICE with the recently proposed semi-supervised learning methods including $\Pi$-Model~\cite{ladder}, Pseudo-Labeling~\cite{pseudo-label}, Mean Teacher~\cite{mean-teacher} MixMatch~\cite{mixmatch}, UDA~\cite{UDA}, ReMinMatch~\cite{remixmatch}, and FixMatch~\cite{fixmatch}, as shown in Table~\ref{table_results_semi}. Here the semi-supervised learning methods use 250 and 1,000 samples with ground truth labels on CIFAR-10 and STL10, respectively. Here the semi-supervised learning results on the commonly used CIFAR-10 and STL10 datasets are from~\cite{fixmatch}.
SPICE uses the same backbone network in FixMatch for fair comparison and were conducted five times for reporting the mean and standard deviation results.
It is found in our experiments that SPICE is comparable to and even better than these state-of-the-art semi-supervised learning methods. Actually, these results demonstrate that SPICE$_\mathrm{s}$ with the reliable pseudo-labeling algorithm can accurately label a set of images without human interaction.

\begin{table}[ht]
\footnotesize
\renewcommand{\arraystretch}{1.3}
\renewcommand\tabcolsep{4pt}
\caption{Feature quality before and after joint training. The format of results for before joint training means the ACC of \emph{supervised / unsupervised / reliable labels} for after joint training means the ACC of  \emph{supervised / unsupervised}. The supervised results were obtained by training a linear classifier while fixing the feature model.}
\label{table_results_joint}
\centering
\begin{tabular}{cccc}

\toprule

Method                          & STL-10  & CIFAR-10 &CIFAR-100-20 \\

\midrule
Before            & 0.908/0.862/0.977   & 0.893/0.845/0.965 & 0.729/0.468/0.677  \\
After             & 0.938/0.920   & 0.922/0.918 & 0.723/0.535   \\

\bottomrule
\end{tabular}
\end{table}

\subsection{Unsupervised Representation Learning}
Here we aim to study the effects of the reliable pseudo-labeling-based joint training in SPICE on representation features. 
As in all unsupervised representation learning methods~\cite{He_2020_CVPR}, the quality of representation features is evaluated by training a linear classifier while freezing the learned feature model using the ground-truth labels.
The results in Table~\ref{table_results_joint} show that the quality of representation features on CIFAR-10 and STL10 is clearly improved after joint training, while keeping similar performance on CIAFR-100-20.
The interpretation is that the feature improvement with joint learning requires very accurate pseudo-labels (the ACC of reliable labels on STL10 and CIFAR-10 are 97.7\% and 96.5\%), so that the improvement is not observed on CIFAR-100-20 (the ACC of reliable labels on CIFAR-100-20 is 67.7\%).
These results indicate that the proposed framework has the potential to improve the unsupervised representation learning by generating accurate and reliable pseudo-labels.

\subsection{Empirical Analysis}

In this subsection, we empirically analyze the effectiveness of different components and options in the proposed SPICE framework. 

\subsubsection{Visualization of cluster semantics}

\begin{figure*}[hbt!]
    \centering
    \includegraphics[width=0.99\textwidth]{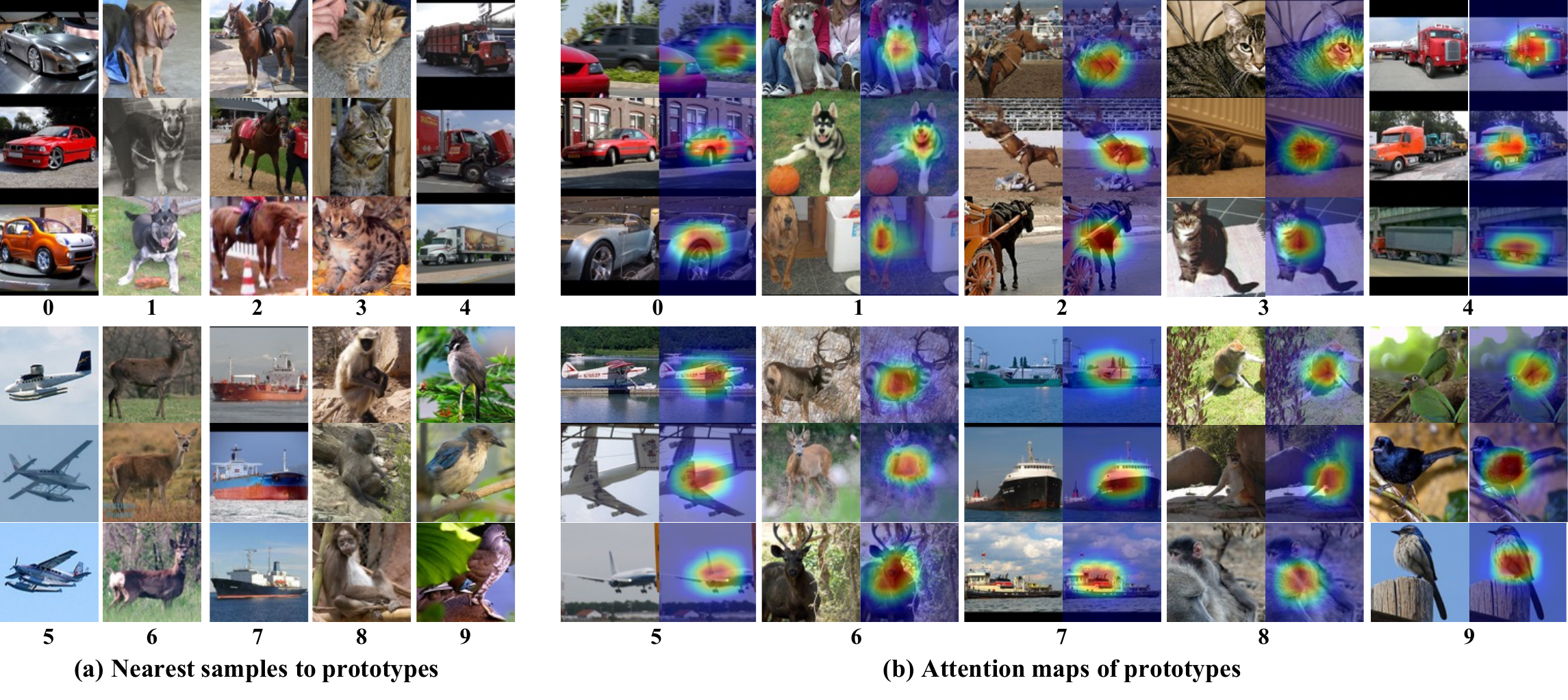}
    \caption{Visualization of learned semantic clusters on STL10. (a) The top three nearest samples to the cluster centers. (b) The attention maps of cluster center on individual images.}
    \label{fig_proto}
\end{figure*}

We visualize the semantic clusters learned by SPICE$_\mathrm{s}$ in terms of the prototype samples and the discriminative regions, as shown in Fig.~\ref{fig_proto}.
Specifically, Fig.~\ref{fig_proto}(a) shows the top three nearest samples of the cluster centers representing the cluster prototypes, and in Fig.~\ref{fig_proto}(b), each cluster includes three samples and each sample is visualized with an original image and an attention map to highlight the discriminative regions. The attention maps of each cluster is computed by computing the cosine similarity between the cluster center (with Eqs.~\eqref{eq_conf} and \eqref{eq_center}) of the whole dataset and the convolutional feature maps of individual images, and then resized and normalized into [0, 1].
It shows that the prototype samples exactly match the human annotations, and the discriminative regions focus on the semantic objects.
For example, the cluster with label `1' captures the `dog' class, and its most discriminative regions exactly capture the dogs at different locations, and similar results can be observed for all other clusters.
The visual results indicate that semantically meaningful clusters are learned, and the cluster center vectors can extract the discriminative features.

\subsubsection{Ablation study}
\label{sec_abl}

\begin{table}[ht]
\footnotesize
\renewcommand{\arraystretch}{1.3}
\renewcommand\tabcolsep{9.8pt}
\caption{ Ablation studies of SPICE$_\mathrm{s}$ on the whole STL10 dataset.}
\label{table_results_abl_whole}
\centering
\begin{tabular}{cccc}

\toprule

Variants                          & ACC&NMI&ARI\\

\midrule
Non-overlap                                      & 0.885$\pm$0.002 & 0.788$\pm$0.003 & 0.771$\pm$0.003     \\
Joint-SH                                      & 0.622$\pm$0.061 & 0.513$\pm$0.037 & 0.437$\pm$0.053     \\
Joint-MH                                      & 0.687$\pm$0.037 & 0.577$\pm$0.029 & 0.512$\pm$0.033     \\
Entropy                                          & 0.907$\pm$0.001 & 0.817$\pm$0.003 & 0.810$\pm$0.003     \\
CE                                               & 0.875$\pm$0.031 & 0.784$\pm$0.017 & 0.764$\pm$0.033     \\
TCE                                              & 0.895$\pm$0.005 & 0.794$\pm$0.010 & 0.787$\pm$0.010     \\
\textbf{SPICE$_\mathrm{s}$}                      & \textbf{0.908$\pm$0.001} & \textbf{0.817$\pm$0.002} & \textbf{0.812$\pm$0.002}    \\
\bottomrule
\end{tabular}
\end{table}

We evaluate the effectiveness of different components of SPICE$_\mathrm{s}$ in an ablation study, as shown in Table~\ref{table_results_abl_whole}. In each experiment, we replaced one component of SPICE$_\mathrm{s}$ with another option, and five trials were conducted to report the mean and standard deviation for each metric.

We first evaluated the effectiveness of the overlap assignment and non-overlap assignment as described in Section~\ref{sec_proto} and Fig.~\ref{fig_sl}.
The results show that the overlap assignment is preferred over the non-overlap assignment, which may be explained by the fact that the non-overlap assignment may introduce extra local inconsistency when assigning the label to a sample far away from the cluster center, as shown in Fig.~\ref{fig_sl} (dashed red circle).

During training SPICE$_\mathrm{s}$, we only optimize the clustering head while freezing the parameters of the feature model.
To demonstrate the effectiveness of this separate training strategy, we compared it with two variants, i.e., jointly training the feature model and a single clustering head (Joint-SH) and jointly training the feature model and multiple clustering heads (Joint-MH). Note that the feature model in the first branch in Fig.~\ref{fig_framework}(b) is still fixed for accurately measuring the similarity.
The results show that the clustering performance for these two variants became significantly worse. On the one hand, the quality of pseudo labels not only depends on the similarity measurement but also the predictions of the clustering head. On the other hand, the performance of the clustering head is also determined by the quality of representation features. When tuning all parameters during training, the feature model tends to be degraded without accurate labels and the clustering head tends to output incorrect predictions in the initial stage, which will harm the pseudo-labeling quality and get trapped in a bad cycle.

Usually, maximizing the entropy over different clusters is the necessary to avoid assign all samples into a single or a few clusters, as demonstrated in GATCluster~\cite{gatcluster}. Thus, we added another entropy loss during training, and the results were not changed. It indicates that our pseudo-labeling process with balance assignment has the ability to prevent trivial solutions. However, when clustering a large number of hierarchical clusters, \eg 200 clusters in Tiny-ImageNet, the entropy loss was found necessary to avoid the empty clusters.

To evaluate the effectiveness of applying double softmax before CE, we replaced it with the plain CE or the CE with a temperature parameter (TCE)~\cite{tce-1}. For TCE, we evaluated different temperature values including 0.01, 0.05, 0.07, 0.1, 0.2, 0.5, 0.8, 2, and found 0.2 achieve the best results that were included for comparison. The results show that TCE is better than CE, which is consistent with the literature~\cite{He_2020_CVPR}.
On the other hand, the results of TCE are inferior to that with the double softmax CE. These experimental results are consistent with the detailed analysis in terms of the derivatives in Appendix~A.


\subsubsection{Clustering head selection}

\begin{figure}[hbt!]
    \centering
    \includegraphics[width=0.45\textwidth]{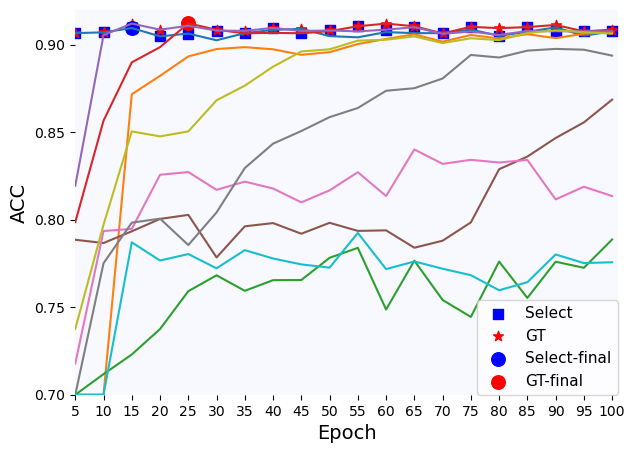}
    \caption{Clustering head selection. Each curve represents the changing process of ACC v.s. epoch of a specific classification head. The blue squares mark the selected best classification head for each epoch, and the red stars represent the corresponding best head evaluated with the ground truth. The blue circle denotes the finally selected head. The red circle is the ground truth best head.}
    \label{fig_select}
\end{figure}

In unsupervised learning, the training process is hard to converge to the best state without the ground truth supervision, such that usually there is a large standard deviation among different trials.
Thus, how to estimate the performance of models in the unsupervised training process to select the potential best model is very important. As introduced in Subsection~\ref{sec_proto}, we use the classification loss defined in Eq.~\eqref{eq_loss} on the whole test dataset to approximate the classification performance; \ie, the smaller loss, the better clustering performance. The model selection process is shown in Fig.~\ref{fig_select}, it can be seen that the performance of the selected clustering head is very close to that of the ground truth selection, proving the effectiveness of this loss metric. In this way, the bad performance clustering heads can be filtered out. The results in Table~\ref{table_results_detail} show that SPICE has a lower standard deviation compared with the competing methods.
Importantly, the light-weight clustering heads can be independently and simultaneously trained without affecting each other and without extra training time.


\subsubsection{Effect of reliable labeling}
\label{sec_cl}

\begin{figure}[hbt!]
    \centering
    \includegraphics[width=0.43\textwidth]{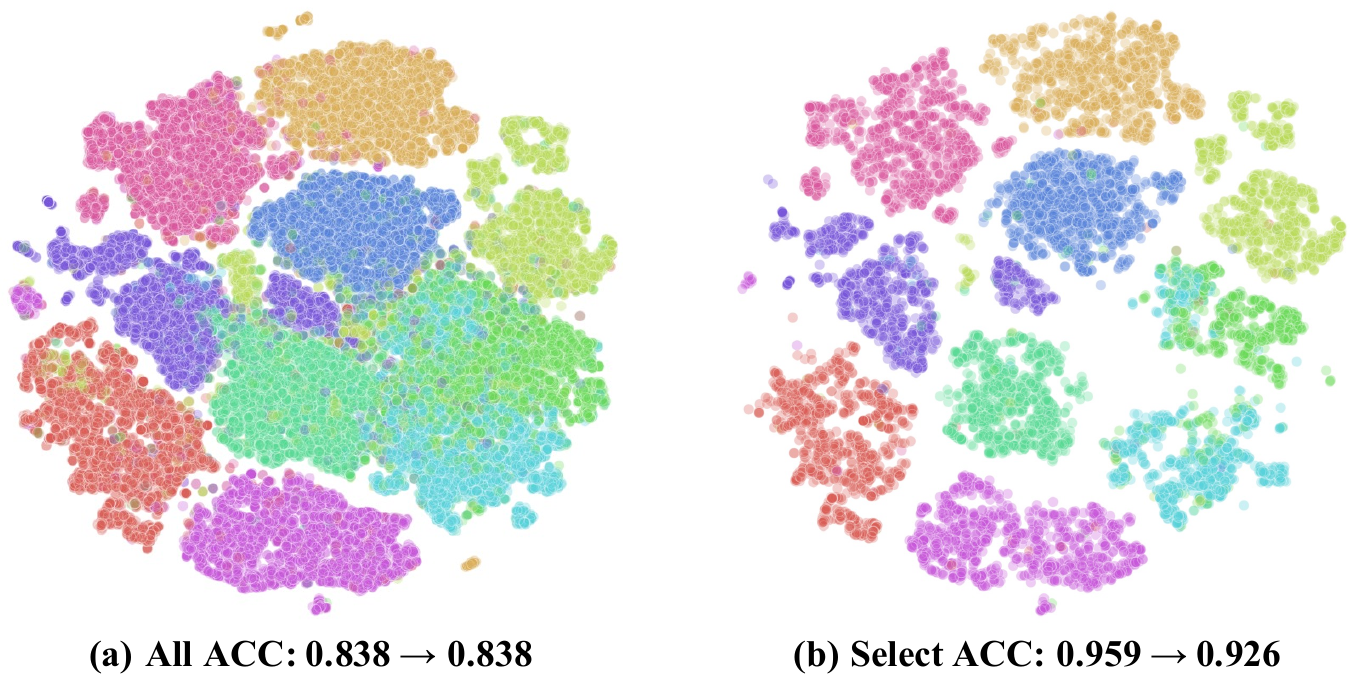}
    \caption{Visualization of reliable labels on CIFAR-10 dataset. Each point denotes a sample in the embedding space, different colors are rendered by the ground-truth labels. (a) The ACC of all samples is 0.838 and the ACC of the jointly trained model using all pseudo-labeled samples is also 0.838.  (b) The ACC of the selected reliable labels is 0.959 and the ACC of the jointly trained model using the reliable samples is 0.926.}
    \label{fig_lc}
\end{figure}

In Section~\ref{sec_semi}, we introduced a reliable pseudo-labeling algorithm to select the reliably labeled images. Here we show the effectiveness of this algorithm in Fig.~\ref{fig_lc}, where t-SNE was used to map the representation features of images in CIFAR-10 to 2D vectors for visualization. In Fig.~\ref{fig_lc}(a), some obvious semantically inconsistent samples are evident, and the predicted ACC of SPICE$_\mathrm{s}$ on all samples is 83.8\%. Using these samples directly for joint training, the ACC is not boosted. Fig.~\ref{fig_lc}(b) shows the selected reliable samples, where the ratio of local inconsistency samples is significantly decreased, and the ACC is increased to 95.9\% correspondingly. Using the reliable samples for joint training, the ACC of SPICE after joint training is significantly boosted (92.6\% v.s. 83.8\%) compared with that of SPICE$_\mathrm{s}$.


\subsubsection{Effect of data augmentation}

\begin{table}[ht]
\footnotesize
\renewcommand{\arraystretch}{1.3}
\renewcommand\tabcolsep{7pt}
\caption{ Results of SPICE$_\mathrm{s}$ with different data augmentation strategies on the whole STL10 dataset, where ResNet34 was used as backbone, and each variant was conducted five times.}
\label{table_results_aug}
\centering
\resizebox{\linewidth}{!}{
\begin{tabular}{llccccccccc}

\toprule

Aug1 & Aug2                          & ACC&NMI&ARI\\

\midrule
Weak   & Weak                         & 0.905$\pm$0.002 & 0.815$\pm$0.003 & 0.808$\pm$0.003     \\
Strong & Weak                         & 0.883$\pm$0.029 & 0.799$\pm$0.019 & 0.781$\pm$0.031     \\
Strong & Strong                       & 0.902$\pm$0.008 & 0.812$\pm$0.009 & 0.803$\pm$0.013     \\
Weak   & Strong                       & \textbf{0.908$\pm$0.001} & \textbf{0.817$\pm$0.002} & \textbf{0.812$\pm$0.002}     \\
\bottomrule
\end{tabular}}
\end{table}

We evaluated the effects of different data augmentations on SPICE$_\mathrm{s}$, as shown in Table~\ref{table_results_aug}, where Aug1 and Aug2 correspond to data augmentations of the second and the third branches in Fig.~\ref{fig_framework}(b). The results show that when the second branch used the weak augmentation and the third branch used the strong augmentation, the model achieved the best performance. Moreover, the model had relatively worse performance when the the second branch in labeling process uses the strong augmentation, which is due to that the labeling process aims to generate reliable pseudo labels that will be compromised by the strong augmentation.
The model performs better when the third branch used the strong augmentation, as it will drive the model to output consistent predictions of different transformations.
Overall, the data augmentation has a small impact on the results, because the pre-trained feature model had been already equipped with the transformation invariance ability.

\section{Discussions}

In this study, the SPICE network significantly improves the image clustering performance over the competing methods and reduces the performance gap between unsupervised and fully-supervised classification. However, there are opportunities for further refinements. First, existing deep clustering methods assume the clustering number $K$ is known. In real applications, we do not always have such a prior. Therefore, how to automatically determine the number of semantically meaningful clusters is an open problem for deep clustering research. Second, to avoid trivial solutions, almost all existing clustering methods assume that the target dataset contains a similar number of samples in each and every cluster, which may or may not be the case in a real-world application. Usually, there are at least two constraints that can be applied to implement this prior, including maximizing the entropy \cite{gatcluster} and balancing assignment \cite{pseudo-label} (and SPICE) that is an optimal solution for maximizing entropy. On the other hand, if we do have a prior distribution of samples as a function of the cluster index, the constraints for these methods can be adapted from the uniform distribution to a specific one.
The ideal clustering method should work well when neither the number of clusters nor the prior distribution of samples per cluster is known, which is a holy grail in this field.
Finally, although the SPICE method achieved the superior results over the existing methods, the progressive training process through the three stages is based on multiple algorithmic ingredients, and could be further unified in an elegant framework that optimizes weak and strong transforms as well, which is beyond the scope of this paper.
Nevertheless, our method is both effective and efficient, and easy to optimize and apply, because each training stage only has a single cross-entropy function and is fairly straightforward.
Importantly, SPICE is an exemplary workflow to synergize both the instance similarity and semantic discrepancy for superior image clustering.

\section{Conclusion}
We have presented a semantic pseudo-labeling framework for image clustering, with the acronym ``SPICE''. 
To accurately measure both the similarity among samples and the discrepancy between clusters for clustering, we divide the clustering network into a feature model and a clustering head, which are first trained separately with the unsupervised representation learning algorithm and the prototype pseudo-labeling algorithm, and then jointly trained with the reliable pseudo-labeling algorithm.
Extensive experiments have demonstrated the superiority of SPICE over the competing methods on six public datasets with an average performance boost of 10\% in terms of adjusted rand index, normalized mutual information, and clustering accuracy.
The SPICE is comparable to or even better than the state-of-the-art semi-supervised learning methods and has the ability to improve the representation features.
Remarkably, SPICE significantly reduces the gap between unsupervised and fully-supervised classification; \eg, only 2\% gap on CIFAR-10.
We believe the basic idea behind SPICE has the potential to help cluster other domain datasets, and apply to other learning tasks.

\ifCLASSOPTIONcompsoc
  \section*{Acknowledgments}
\else
  \section*{Acknowledgment}
\fi
This work was supported in part by NIH/NCI under Award numbers R01CA237267, and in part by NIH/NHLBI under Award number R01HL151561.


\appendices

\renewcommand\thefigure{\thesection.\arabic{figure}}    
\renewcommand{\theequation}{\thesection-\arabic{equation}}    

\setcounter{figure}{0}    
\setcounter{equation}{0}

\section{Analysis of the double softmax mechanism for pseudo-labeling-based image clustering}
\label{ap_ds}
In this appendix, we give a detailed analysis on the double softmax mechanism for training clustering head based on the pseudo-labels.
Considering that the pseudo-labels are not as accurate as the ground truth labels, the basic idea behind the double softmax implementation is to reduce the learning speed especially when the predictions are of low probabilities.
Next, we show how the double softmax mechanism affects the gradients relative to the normal single softmax.
Specifically, we denote the output score of the fully-connected layer as $\vct{a}=[a_1, a_2, \ldots, a_K]\T$, where $K$ represents the number of clusters.
Then, the output  of the first softmax function is
\begin{align}
    p_k = \frac{\exp(a_k)}{\sum_{j=1}^K \exp(a_j)},
\end{align}
and the output of the second softmax function is
\begin{align}
    p'_k = \frac{\exp(p_k)}{\sum_{j=1}^K \exp(p_j)}.
\end{align}

Considering the following cross entropy loss
\begin{align}
    \mathcal{L}(\vct{y}, \vct{p}) = -\sum_{k=1}^K y_{k} \log(p_k),
\end{align}
where $y_k \in \{0, 1\}$ is the pseudo label, the derivative of cross-entropy loss with respect to $a_k$ is:
\begin{itemize}
\item When we use the single softmax, the derivative of $\mathcal{L}(\vct{y}, \vct{p})$ with respect to $a_k$ is
\begin{align}
    \label{eq_a1}
    \frac{\partial \mathcal{L}(\vct{y}, \vct{p})}{\partial a_k} = p_k - y_k.
\end{align}

\item When we use the double softmax composition, the derivative of $\mathcal{L}(\vct{y}, \vct{p}')$ with respect to $a_k$ is
\begin{align}
    \label{eq_a2}
    \frac{\partial \mathcal{L}(\vct{y}, \vct{p}')}{\partial a_k} &= \sum_{j=1}^{K} \frac{\partial \mathcal{L}(\vct{y}, \vct{p}')}{\partial p_j} \frac{\partial p_j}{\partial a_k} =\sum_{j=1}^{K}(p'_j-y_j)p_j(\delta_{jk}-p_k),
\end{align}
where $\delta_{jk}=1$ if $k=j$, else $\delta_{jk}=0$.
\end{itemize}

Then, we compute the ratio $r_k = \frac{\partial \mathcal{L}(\vct{y}, \vct{p}')}{\partial a_k} / \frac{\partial \mathcal{L}(\vct{y}, \vct{p})}{\partial a_k}$ as in Eq. (\ref{eq_a4}):
\begin{align}
    \label{eq_a4}
    r_k = \frac{\sum_{j=1}^{K}(p'_j-y_j)p_j(\delta_{jk}-p_k)}{p_k - y_k}
\end{align}

During training, we know the pseudo cluster label is $c$, and $y_k = 1$ if $k = c$, else $y_k = 0$.
\begin{itemize}
\item If $k=c$, we have
\begin{align}
    \label{eq_a5}
    r_c &= \frac{\sum_{j=1}^{K}(p'_j-1)p_j(\delta_{jc}-p_c)}{p_c - 1} \\
    \label{eq_a6}
        & = \sum_{j=1}^K (1-p'_j)p_j - \sum_{j=1, j\ne c}^K \frac{(1-p'_j)p_j}{1-p_c} \\
        \label{eq_a7}
        & = (1-p'_c)p_c + \frac{p_c}{1-p_c} \sum_{j=1, j\ne c}^K (1-p'_j)p_j.
\end{align}
According to Eqs.~\eqref{eq_a6} ($r_c < 1$) and~\eqref{eq_a7} ($r_c > 0$), we have $r_c \in (0, 1)$.

\item If $k\ne c$, then
\begin{align}
    \label{eq_a8}
    r_k &= \frac{\sum_{j=1}^{K}(p'_j-0)p_j(\delta_{jk}-p_k)}{p_k - 0} \\
    \label{eq_a9}
        & = p'_k - \sum_{j=1}^K p'_j p_j 
\end{align}
According to Eq.~\eqref{eq_a9}, we have $r_k \propto p'_k \propto p_k$ as the second term is a constant for a specific prediction $\vct{p}$, note that here the index $k$ is the variable. 
\end{itemize}
Thus, the overall learning speed for the double softmax is adaptively reduced by a foctor of $r_k < 1$ compared with that for the normal single softmax. Importantly, when $k \ne c$, the learning speed for smaller $p_k$ is relatively slower than that for larger probabilities.
In this way, the double softmax implementation will prevent the low probabilities from being too small after the optimization with current pseudo-labels, which will benefit the training process given imperfect pseudo-labels.
In another angle, the relatively larger learning speed for the false high probability will accelerate the sample changing between different clusters during training, which will benefit the dynamically searching process for good clusters.

\begin{figure*}[hbt!]
    \centering
    \includegraphics[width=0.99\textwidth]{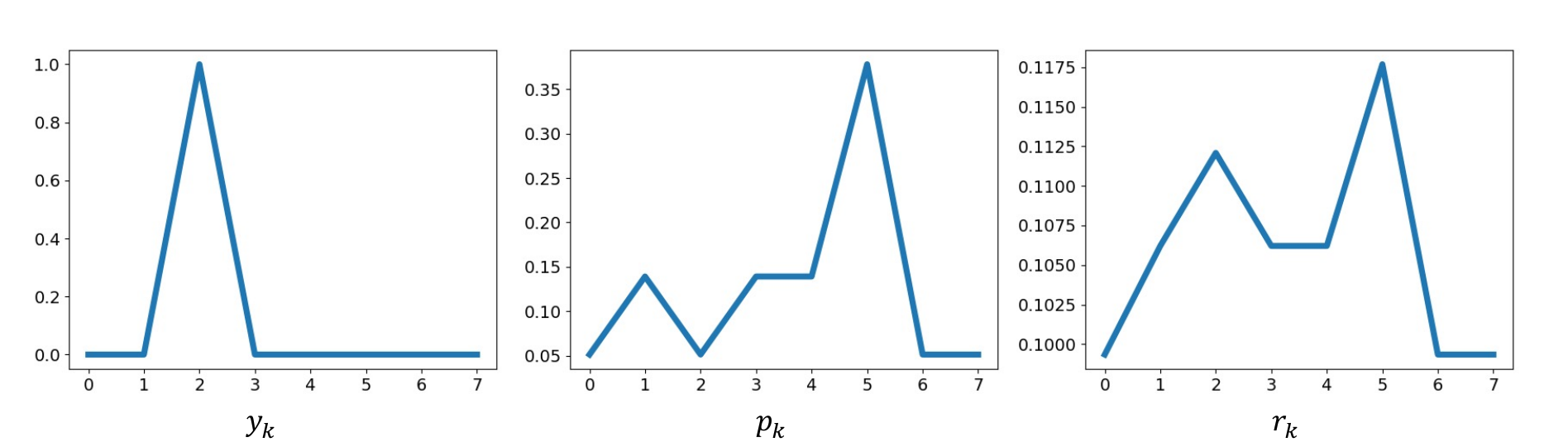}
    \caption{Simulation results for double softmax mechanism.}
    \label{fig_dsce}
\end{figure*}

The corresponding numerical simulation results are shown in Fig. \ref{fig_dsce}, where there are $K=8$ clusters and the pseudo-label is $c=2$.
We can see that, except $k=2$, the larger predicted probability $p_k$ corresponds to the relatively larger ratio $r_k$. For $k=2$, although $p_k$ is small, $r_k$ is relatively large as there is a big difference between the prediction and the pseudo-label.
For the false high-probability, i.e. $k=5$, $r_k$ is the largest among other ratios. These simulation results are consistent with the above analytical results.

\ifCLASSOPTIONcaptionsoff
  \newpage
\fi



%
\bibliographystyle{IEEEtran}
\bibliography{IEEEabrv,./spice}

%








\end{document}